\documentclass[letterpaper]{article} % DO NOT CHANGE THIS
\usepackage{aaai25}  % DO NOT CHANGE THIS
\usepackage{times}  % DO NOT CHANGE THIS
\usepackage{helvet}  % DO NOT CHANGE THIS
\usepackage{courier}  % DO NOT CHANGE THIS
\usepackage[hyphens]{url}  % DO NOT CHANGE THIS
\usepackage{graphicx} % DO NOT CHANGE THIS
\usepackage{natbib}  % DO NOT CHANGE THIS AND DO NOT ADD ANY OPTIONS TO IT
\usepackage{caption} % DO NOT CHANGE THIS AND DO NOT ADD ANY OPTIONS TO IT
\usepackage{algorithm}
\usepackage{algorithmic}
\usepackage{marvosym}

\usepackage{color}
\usepackage{subfigure}
\usepackage{multirow}
\usepackage{multicol}
\usepackage{amsmath}
\usepackage{amssymb}
\usepackage{dsfont}
\usepackage{booktabs}
\usepackage[urlcolor=blue]{hyperref}
\usepackage{newfloat}
\usepackage{listings}
\DeclareCaptionStyle{ruled}{labelfont=normalfont,labelsep=colon,strut=off} % DO NOT CHANGE THIS
\floatstyle{ruled}
\newfloat{listing}{tb}{lst}{}
\floatname{listing}{Listing}
\title{AlphaFolding: 4D Diffusion for Dynamic Protein Structure Prediction with Reference and Motion Guidance}
\author{
    Kaihui Cheng\textsuperscript{\rm 1}\equalcontrib, 
    Ce Liu\textsuperscript{\rm 2}\equalcontrib, 
    Qingkun Su\textsuperscript{\rm 2}, 
    Jun Wang\textsuperscript{\rm 2}, 
    Liwei Zhang\textsuperscript{\rm 2}, 
    Yining Tang\textsuperscript{\rm 1}, 
    Yao Yao\textsuperscript{\rm 3}, \\
    Siyu Zhu\textsuperscript{\rm 1,2 \Letter}, 
    Yuan Qi\textsuperscript{\rm 1,2 \Letter}
}
\affiliations{
    \textsuperscript{\rm 1}Fudan University\\ 
    \textsuperscript{\rm 2}Shanghai Academy of Artificial Intelligence for Science\\
    \textsuperscript{\rm 3}Nanjing University\\
}

\usepackage{bibentry}
\begin{document}

\maketitle

\begin{abstract}
Protein structure prediction is pivotal for understanding the structure-function relationship of proteins, advancing biological research, and facilitating pharmaceutical development and experimental design.
While deep learning methods and the expanded availability of experimental 3D protein structures have accelerated structure prediction, the dynamic nature of protein structures has received limited attention.
This study introduces an innovative 4D diffusion model incorporating molecular dynamics (MD) simulation data to learn dynamic protein structures.
Our approach is distinguished by the following components: 
(1) a unified diffusion model capable of generating dynamic protein structures, including both the backbone and side chains, utilizing atomic grouping and side-chain dihedral angle predictions; 
(2) a reference network that enhances structural consistency by integrating the latent embeddings of the initial 3D protein structures; 
and (3) a motion alignment module aimed at improving temporal structural coherence across multiple time steps.
To our knowledge, this is the first diffusion-based model aimed at predicting protein trajectories across multiple time steps simultaneously.
Validation on benchmark datasets demonstrates that our model exhibits high accuracy in predicting dynamic 3D structures of proteins containing up to 256 amino acids over 32 time steps, effectively capturing both local flexibility in stable states and significant conformational changes. \url{https://fudan-generative-vision.github.io/AlphaFolding/#/}
\end{abstract}

\begin{figure*}[t]
\centering
\includegraphics[width=\linewidth]{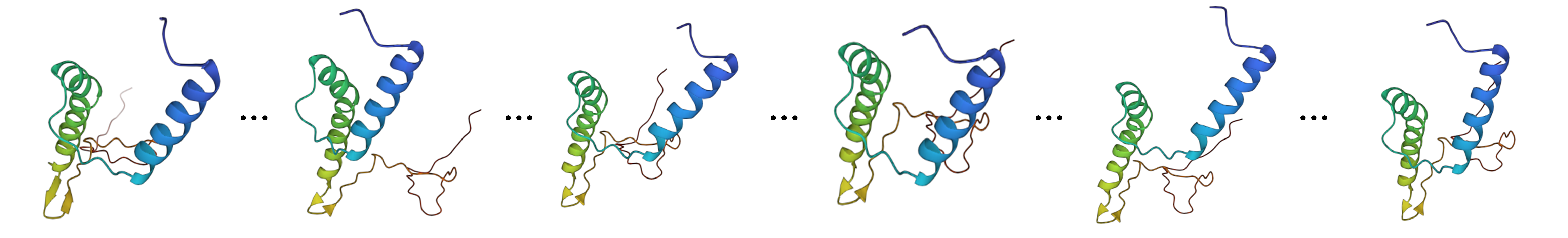}
\caption{\small \textbf{4D dynamic protein prediction.} Given an initial 3D structure for reference, our model predicts dynamic proteins at the following 32 time steps simultaneously. We present the predicted 3D protein structures at the intermediate time steps for illustration. }
\label{fig:teaser}
\vspace{-4mm}
\end{figure*}

\section{Introduction}
The observation and prediction of protein structures are pivotal in elucidating the complex relationship between protein conformation and function.
This understanding drives significant advancements in biological research and pharmaceutical development, while also providing essential guidance for related experimental endeavors and design strategies.
The 3D architecture of a protein is intricately encoded within its linear 1D amino acid sequence, which fundamentally dictates the protein's biological functionality.
Deciphering the process of protein folding has long posed a formidable challenge within the domain of computational biophysics. 
Key challenges in protein structure prediction include the accurate identification of suitable templates for protein structures, particularly for sequences lacking closely related templates; the refinement of these templates to closely resemble the native state; the enhancement of force field precision and conformational exploration; as well as the effective management of computational costs associated with predicting protein structures. 
This is especially pertinent in scenarios involving free modeling, where structures must be generated de novo.

Recent advancements in deep learning techniques, coupled with the exponential growth of experimental protein structures within the Protein Data Bank (PDB)~\cite{bank1971protein}
% , which now houses over 360,000 three-dimensional structures spanning 21 proteomes, 
have markedly propelled learning-based structural studies.
AlphaFold2~\cite{jumper2021highly} has introduced a groundbreaking approach to predicting 3D protein structures, achieving accuracy comparable to experimental methods.
In tandem, RoseTTAFold~\cite{baek2021rose} has enhanced predictive capabilities by incorporating a three-track network architecture, resulting in superior accuracy.
Concurrently, ESMFold~\cite{rives2019biological} and OmegaFold~\cite{wu2022omega} capitalize on high-capacity transformer language models trained on evolutionary data to derive unsupervised representations of protein sequences.
Moreover, the accessibility of large-scale data repositories has substantially advanced research in protein conformation sampling, which seeks to generate diverse structural conformations. 
For instance, Distributional Graphformer (DiG) facilitates the prediction of equilibrium distributions in molecular systems, enabling efficient generation of diverse conformations and the estimation of state densities~\cite{zheng2024predicting}.
EigenFold~\cite{jing2023eigenfold} approaches protein structures as systems of harmonic oscillators, fostering a cascading-resolution generative process along the system's eigenmodes.
AlphaFlow~\cite{jing2023alphaflow} optimizes single-state predictors through a custom flow matching framework to develop sequence-conditioned generative models of protein architectures. 
Building on its predecessor, AlphaFold3~\cite{abramson2024accurate} utilizes a diffusion network and updated algorithmic architecture to incorporate joint structures across proteins, nucleic acids, small molecules, ions, and modified residues.
Furthermore, Str2Str~\cite{lu2024strstr} introduces an innovative framework for structure-to-structure translation, capable of zero-shot conformation sampling while maintaining roto-translation equivariance. ConfDiff~\cite{wang2024protein} further  leverage force guidance for rich diversity and high fidelity.
Despite these significant advancements in structural and conformational predictions, the exploration of dynamic protein structures remains underdeveloped. 
This study aims to address this gap, focusing on the dynamic aspects of protein structures.

Molecular dynamics (MD) simulations serve as crucial tools in the fields of computational biology, biophysics, and chemistry, providing a comprehensive and dynamic perspective of molecular systems. 
These simulations generate substantial high-quality data, which can be exploited for data-driven, learning-based methodologies. 
Nevertheless, the computational expense associated with MD simulations typically scales cubically with the number of electronic degrees of freedom. 
Moreover, critical biomolecular processes, such as conformational changes, often occur on timescales that surpass the capabilities of classical all-atom MD simulations. 
In response, deep learning techniques have been employed to address these limitations. 
Despite these advancements, existing methods are predominantly applicable to proteins with significantly fewer atoms than typical proteins, necessitating the adoption of coarse-grained atomic representations for larger systems. 
This study aims to leverage extensive, high-quality MD data to generate dynamic structures of proteins comprising up to hundreds of amino acids, including complex structures with complete side-chain representations. 
Our approach seeks to extend the applicability of MD simulations to larger and more intricate protein systems, thereby enhancing our understanding of their dynamic behaviors.

This paper presents an innovative approach to modeling dynamic protein structures utilizing a 4D diffusion model. 
Our research is concentrated on three primary areas:
Firstly, we propose a unified diffusion model designed to predict protein structures that encompass both backbone and side-chain components. 
By organizing atoms within each residue into rigid groups to minimize the degrees of freedom, our framework efficiently simulates protein motion for structures with hundreds of residues. 
The amino acid sequence is represented by node and edge features derived from structure prediction models, which guide the diffusion model for precise protein generation. 
Unlike methods constrained to \textit{de novo} structure prediction, we incorporate side-chain dihedral angle predictions and introduce an amino acid atomic model to accurately recover individual atomic coordinates based on dihedral angles.
Secondly, the initial 3D protein structure is integrated as a condition and encoded through a reference network for latent embedding, thereby incorporating relevant features into the denoising diffusion network. 
The reference network is instrumental in maintaining structural consistency of proteins during motion.
Thirdly, we propose a motion alignment module within the score-based diffusion network, which includes temporal attention layers to aggregate kinetic information from adjacent frames within the diffusion model. 
This enhancement improves the coherence of motion in generated dynamic proteins, mitigating abrupt transitions during motion. 
Thus, our diffusion model effectively generates dynamic protein structures across multiple time steps simultaneously, enhancing efficiency and ensuring the prediction of consistent sequences of protein structures within a temporal framework.
In summary, our approach enhances the efficacy of dynamic protein structure generation while ensuring the prediction of coherent and temporally consistent sequences.

In this investigation, we conducted a comprehensive qualitative and quantitative analysis utilizing widely recognized benchmark datasets, including ATLAS~\cite{vander2024atlas} and Fast-Folding~\cite{kresten2011@how} protein datasets. 
Our study successfully achieved dynamic protein structure predictions for sequences of up to 256 amino acids across 32 time steps. 
This capability enabled us to model dynamic protein conformations sampled at various temporal intervals, demonstrating notable accuracy in capturing both subtle intra-conformational motions and significant inter-conformational changes. 
The findings of this research represent a significant advancement in the field of dynamic protein structure prediction, contributing valuable insights for future developments in this domain.

\begin{figure*}[t]
  \centering
  \includegraphics[width=1.0\linewidth]{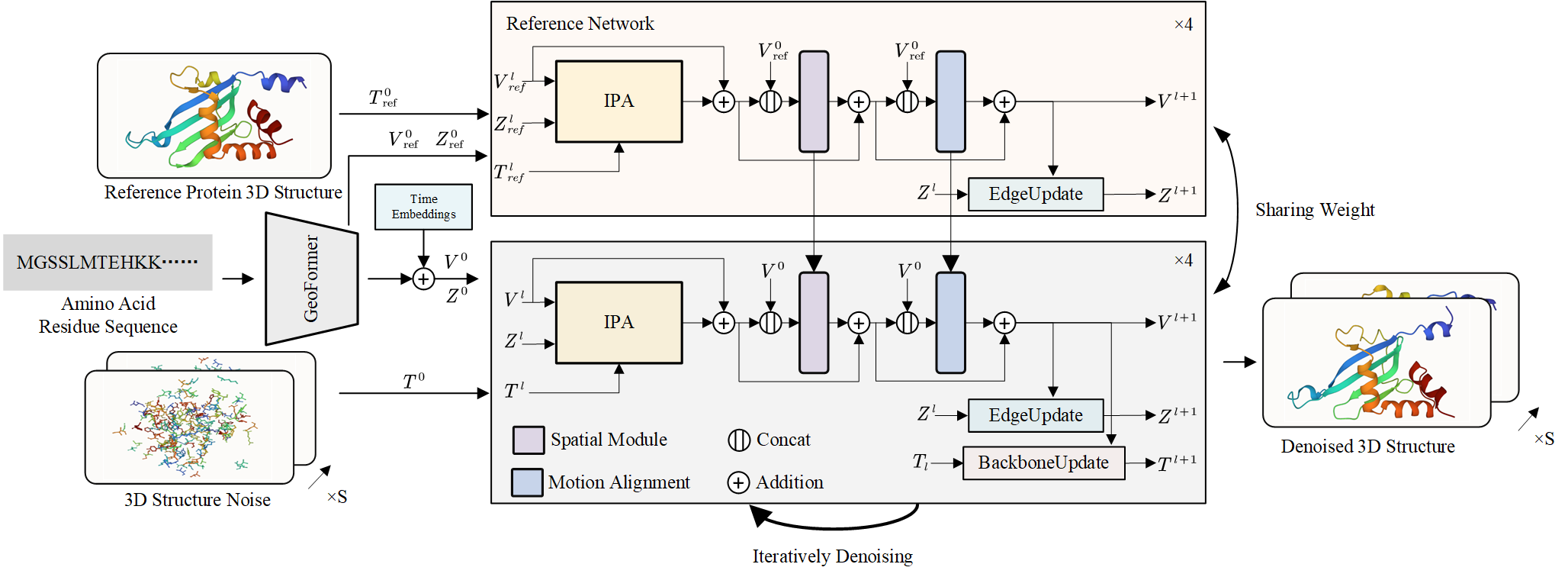}
  \caption{\textbf{The overview of our proposed approach.}
  The diffusion-based generative model that takes the reference structure and corresponding residue sequence as input and produces a sequence of denoised 3D protein structures as output.
  We use the 3D structure embedder and GeoFormer for embedding the 3D protein structures and residue sequences, respectively.
  The Invariant Point Attention (IPA) updates node features by integrating information from the explicit frames of residues. 
  The Reference Network and Motion Alignment module are based on the reference 3D protein structure to capture a sequence of 3D protein dynamics. 
  The entire generative model is formulated as a score-based diffusion model, with node and edge feature embedding updated through the EdgeUpdate and BackboneUpdate modules. 
  }
  \label{fig:network}
  \vspace{-4mm}
\end{figure*}

\section{Related Work}
\paragraph{\textit{De Novo} Protein Design.}
The task of \textit{de novo} protein design~\cite{ trippe2023diffusion, luo2022anti, anand2022protein} involves generating novel proteins based on physical principles, with specified structural and/or functional properties.
FoldFlow~\cite{bose2023se} introduces a simulation-free approach for learning deterministic continuous-time dynamics and matching invariant target distributions on $\mathrm{SE}(3)$. 
VFN-Diff~\cite{mao2024de} presents the Vector Field Network (VFN), which enables network layers to perform learnable vector computations between coordinates of frame-anchored virtual atoms, thereby enhancing the capability for modeling frames.
In recent years, with the rapid development of diffusion-based generative models~\cite{ho2020denoising,song2020score,zhu2024champ,xu2024hallo}, these technologies have also been applied to \textit{de novo} protein design.
RFDiffusion~\cite{watson2023novo} fine-tunes the RoseTTAFold~\cite{baek2021rose} structure prediction network on protein structure denoising tasks, resulting in a generative model of protein backbones based on the diffusion model in the formulation of DDPM~\cite{ho2020denoising,nichol2021improved}. 
$\mathrm{SE}(3)$-Diff~\cite{yim23se3} establishes the theoretical foundations of $\mathrm{SE}(3)$ invariant diffusion models across multiple frames, facilitating the learning of $\mathrm{SE}(3)$ equivariant scores over multiple frames using a score-based diffusion model~\cite{song2020score,song2020improved}.
In this paper, we follow the score-based diffusion model~\cite{song2020score,song2020improved}, extending it not only to protein structure prediction but also to the dynamic motion within the temporal domain.

\paragraph{Prediction of 3D Protein Structure.}
Predicting the 3D structure of proteins from their amino acid sequences has long been a significant challenge in biology. 
Various approaches, including thermodynamic and kinetic simulations and bioinformatics analyses, have been proposed. 
This paper focuses on deep learning-based methods.
An early deep learning effort, Raptor-X~\cite{xu2019raptor}, utilizes a dilated ResNet to predict atom pair distances. 
Subsequently, trRosetta~\cite{yang2020rr} enhances accuracy by predicting inter-residue geometries. 
AlphaFold2~\cite{jumper2021highly} marks a milestone with its novel attention mechanisms and training procedures, leveraging evolutionary, physical, and geometric constraints to significantly improve accuracy. 
RoseTTAFold~\cite{baek2021rose} further refines network architectures with a three-track network, achieving superior accuracy.
Additionally, ESMFold~\cite{rives2019biological} and OmegaFold~\cite{wu2022omega} employ high-capacity transformer language models trained on evolutionary data in an unsupervised manner to learn protein sequence representations. 
Recently, AlphaFold3~\cite{abramson2024accurate} extends protein structure prediction using a diffusion network and an updated algorithmic architecture, encompassing joint structures of proteins, nucleic acids, small molecules, ions, and modified residues.
However, the aforementioned works primarily focus on static structure prediction using diffusion generative models. 
In contrast, this paper addresses the prediction of dynamic structures over temporal sequences.

\paragraph{Protein Conformation Sampling.}
Proteins are dynamic macromolecules, where conformational changes play critical roles in biological processes.
To obtain a diverse set of conformations, classical approaches such as MSA subsampling have been employed, which subsample the Multiple Sequence Alignment (MSA) input to AlphaFold2.
Recently, diffusion models have emerged for protein conformation generation. Distributional Graphformer (DiG) predicts the equilibrium distribution of molecular systems, enabling efficient generation of diverse conformations and estimation of state densities~\cite{zheng2024predicting}.
EigenFold~\cite{jing2023eigenfold} models the structure as a system of harmonic oscillators, naturally inducing a cascading-resolution generative process along the eigenmodes of the system.
AlphaFlow~\cite{jing2023alphaflow} fine-tunes single-state predictors under a custom flow matching framework to obtain sequence-conditioned generative models of protein structure
Str2Str~\cite{lu2024strstr} adopts a novel structure-to-structure translation framework capable of zero-shot conformation sampling with roto-translation equivariant properties. 
ConfDiff~\cite{wang2024protein} incorporates a force-guided network with score-based diffusion models to generate diverse conformations while preserving high fidelity.
It is important to note that protein conformation sampling predicts the distribution of structures rather than structures within the temporal domain.

\paragraph{Learning Based Molecular Dynamics.}
Deep learning has significantly impacted complex atomic systems by reducing the need for time-consuming calculations~\cite{noe2020machine, merchant2023scaling, kearnes2016molecular, pfau2020ab}. 
Applications include estimating free energy surfaces~\cite{behler2007generalized}, constructing Markov state models of molecular kinetics~\cite{mardt2017vamp}, and generating samples from equilibrium distributions~\cite{jing2023alphaflow}.
Here, we briefly review research on learning kinetics models.
VAMPNet~\cite{mardt2017vamp} introduces a variational approach for Markov processes (VAMP) to develop a deep learning framework for molecular kinetics.
DiffMD~\cite{fang2023diffmd} employs a diffusion model to estimate the gradient of the log density of molecular conformations. 
DFF~\cite{arts2023two} leverages connections between score-based generative models, force fields, and molecular dynamics to learn a coarse-grained force field without requiring force inputs during training.
However, these approximations are designed for general purposes and make limited use of prior knowledge of proteins. 
Consequently, learning atomic interactions incurs high computational costs, restricting their application to large molecules.
In this paper, the objective is to generate dynamic 3D structures of proteins encompassing hundreds of amino acids across numerous time steps.

\section{Preliminaries}\label{sec:preliminary}
\paragraph{Protein Parameterization.}
We adopt the frame-based representation of protein structure used in AlphaFold2 and extend it to incorporate a temporal dimension accounting for structural changes over time.
A static protein comprises a sequence of amino acid residues, each parameterized by a backbone frame, consisting of atoms [$\mathtt{N}, \mathtt{C}_{\alpha}, \mathtt{C}$] with $\mathtt{C}_{\alpha}$ positioned at the origin $(0, 0, 0)$. 
We hence define a dynamic protein composed of $N$ amino acid residues, each parameterized by a backbone frame that undergoes transformations across $S$ time steps. 
Those frames are transformed by special Euclidean transformations that preserve orientations from the local frames to a global reference frame, represented by $T_{s,i}=[R_{s,i}, X_{s,i}]\in\mathrm{SE}(3)$, where $s\in\{1,...,S\}$, $i\in\{1,...,N\}$, $R_{s,i}\in\mathrm{SO}(3)$ is a $3\times3$ rotation matrix, and $X_{s,i}\in\mathbb{R}^3$ is the translation vector.
All additional atoms coordinates in a residue are organized into rigid groups based on their dependency on torsion angles, such that all atoms within a rigid group maintain constant relative positions and orientations to preserve the chemical integrity of the structure. 
This setup allows each residue to be parameterized by torsion angles $\alpha_{s,i}\in\mathbb{R}^7$ that model the rotations required to align atom groups relative to the backbone.
The angles facilitate the precise adjustment of atom positions within each frame, and the transformation parameters allow the model to reconstruct all atom positions from idealized, experimentally determined coordinates over time.

\paragraph{Score-based Modeling on $\mathrm{SE}(3)^{S\times N}$.}
The score-based model functions by diffusing a data distribution towards a noise distribution through a stochastic differential equation (SDE) and then learning to reverse this diffusion to generate samples. 
This process entails systematically reducing the structure in the data by introducing noise until the original signal is almost entirely removed. 
In our study, we diffuse the frames $T=[T_{s,i}]\in\mathrm{SE}(3)^{S\times N}$ following the prior work~\cite{yim23se3}.
More specifically, we construct two independent forward processes for $R=[R_{s,i}]\in\mathrm{SO}(3)^{S\times N}$ and $X=[X_{s,i}]\in\mathbb{R}^{S\times N\times 3}$ respectively: 
\begin{align}
\mathrm{d}T^{(t)} &= [\mathrm{d}R^{(t)}, \mathrm{d}X^{(t)}] \notag\\
                  &= \left[0, -\frac{1}{2}X^{(t)}\right]\mathrm{d}t + [\mathrm{d}B^{(t)}_{\mathrm{SO}(3)^{S\times N}}, \mathrm{d}B^{(t)}_{\mathbb{R}^{S \times N \times 3}}],
\end{align}
where $B^{(t)}_{\mathrm{SO}(3)^{S\times N}}$ and $B^{(t)}_{\mathbb{R}^{S\times N\times 3}}$ are the Brownian motion on $\mathrm{SO}(3)^{S\times N}$ and $\mathbb{R}^{S\times N\times 3}$ respectively, and $t\in[0, 1]$ denotes the diffusion time variable. 
Superscripts in parentheses are used to represent specific time step. 
Lowercase letters denote deterministic variables, and uppercase letters denote random variables.

Accordingly, the associated backward process is given by the equation $\mathrm{d}\overleftarrow{T}^{(t)}=[\mathrm{d}\overleftarrow{R}^{(t)}, \mathrm{d}\overleftarrow{X}^{(t)}]$, where
\begin{align}
    \mathrm{d}\overleftarrow{R}^{(t)} &= \nabla\log p_{1-t}(\overleftarrow{R}^{(t)})\mathrm{d}t + \mathrm{d}B^{(t)}_{\mathrm{SO}(3)^{S\times N}}, \\ 
    \mathrm{d}\overleftarrow{X}^{(t)} &= (\frac{1}{2}\overleftarrow{X}^{(t)}+\nabla\log p_{1-t}(\overleftarrow{X}^{(t)}))\mathrm{d}t + \mathrm{d}B^{(t)}_{\mathbb{R}^{S\times N\times 3}}.
\end{align}
Then, we can learn the score
\begin{equation}
\nabla \log p_t(T^{(t)})=[\nabla \log p_t(R^{(t)}), \nabla \log p_t(X^{(t)})]
\end{equation}
with neural networks $s_{\theta}(t, T^{(t)})$ trained by minimizing the denoising score matching loss:
\begin{equation}
    \mathcal{L}(\theta) = \mathbb{E}[\lambda_t||\nabla \log p_{t|0}(T^{(t)}|T^{(0)})-s_{\theta}(t, T^{(t)})||^2],
    \label{eq:dsm}
\end{equation}
where $\lambda_t\in\mathbb{R}^+$ is a weight, the expectation is taken over $t\sim\mathcal{U}[0,1]$.
and
\begin{equation}
\nabla \log p_{t|0}(T^{(t)}|T^{(0)}) = \begin{aligned}[t]
&[\nabla \log p_{t|0}(R^{(t)}|R^{(0)}), \\
&\nabla \log p_{t|0}(X^{(t)}|X^{(0)})].
\end{aligned}
\end{equation}
\section{Methodology}

The proposed methodology requires as input a sequence of amino acid residues, the reference 3D structure of a protein at a specific time step, and, the 3D structures of additional proteins from preceding time steps;
and the output is the predicted protein trajectories for subsequent time steps. 
The paper commences with an overview of the generative network in Section~\ref{subsec:network}. 
In Section~\ref{subsec:motion}, we present the proposed reference network and the motion alignment approach for learning temporal dynamic structures. 
Furthermore, Section~\ref{subsec:loss} discusses the loss function employed, while Section~\ref{subsec:train} provides detailed information regarding the training and inference processes.

\begin{figure*}[t]
\centering
\subfigure[Spatial Module]{
\begin{minipage}[b]{0.53\textwidth}
\includegraphics[width=1.0\textwidth]{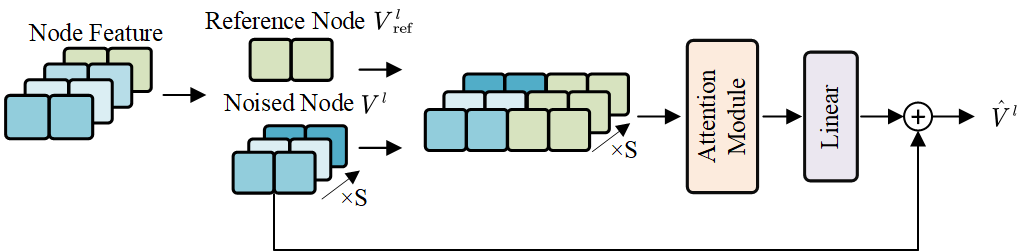}
\label{fig:spatial}
\end{minipage}
}
\subfigure[Motion Alignment]{
\begin{minipage}[b]{0.43\textwidth}
\includegraphics[width=1.0\textwidth]{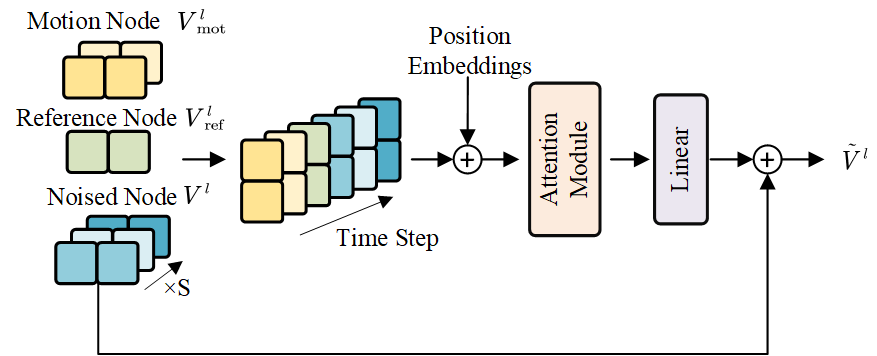}
\label{fig:temporal}
\end{minipage}
}
\caption{\textbf{Structure of spatial module and motion alignment.}
              The spatial module encodes the structural characteristics of the reference 3D protein structure from reference network to preserve its features during dynamic structure generation. 
              The motion alignment is comprised of stacked temporal transformer layers and used to generate protein dynamics. The depth of the color indicates different time steps.}
\label{fig:spatial_temporal}
\vspace{-4mm}
\end{figure*}

\subsection{Network Overview}\label{subsec:network}
The architecture of our model is depicted in Figure~\ref{fig:network}. To capture the dynamic behavior of a protein composed of $N$ residues across $S$ time steps, we utilize node features $V^l=[V^l_{s,i}]\in\mathbb{R}^{S\times N \times D_V}$ and edge features $Z^l=[Z^l_{s,(i,j)}]\in\mathbb{R}^{S\times N\times N\times D_Z}$. $V^l_{s,i}$ denotes the feature of the residue $i$ at the time step $s$ in layer $l$, and $Z^l_{s,(i,j)}$ encodes the relationship between residues $i$ and $j$ at the time step $s$  in layer $l$.
Positional attributes of atoms are represented by frames $T_{s,i}$ and torsion angles $\alpha_{s,i}$.
The reference structure is defined by $V^{l}_{\texttt{ref}}$, $Z^{l}_{\texttt{ref}}$, and $T^{l}_{\texttt{ref}}$. 
Motion structures are characterized by $V^{l}_{\texttt{mot}}$, $Z^{l}_{\texttt{mot}}$, and $T^{l}_{\texttt{mot}}$, which describe multi-order motion information, including velocity, acceleration, and other related parameters across $M$ time steps.

\paragraph{Feature Embedding of Amino Acid Sequence.}
For a given amino acid sequence, we initially extract node and edge features using the GeoFormer protein prediction method~\cite{wu2022omega}. 
These features are further enriched by encoding the diffusion time step, resulting in initial features $V^0_{s,i} \in\mathbb{R}^{N\times D_V}$  and $Z^0_{s,(i,j)} \in\mathbb{R}^{N\times N\times D_Z}$ for each residue $i$ and each pair of residues $(i,j)$. 
The noisy 3D structures $T^0_{s,i}$ are sampled from the Isotropic Gaussian on SO(3) and the Gaussian distribution for capturing rotation and translation.

\paragraph{Invariant Point Attention.}\label{sec:ipa}
We apply IPA mechanism in our networks in each layer $l$, it utilizes node features $V^l_{s,i}$, edge features $Z^l_{s,(i,j)}$, and frames $T^l_{s,i}$ as inputs.
Each node feature $V^l_{s,i}$ generates query, key, and value points, which are subsequently transformed using the frame $T^l_{s,i}$.
A self-attention mechanism aggregates these points based on attention scores at each time step $s$, integrating edge feature information and producing updated node features  through fully-connected layers. %$V^{l+1}_{s,i}$
To preserve reference coordinates, we introduce features without implementing the mapping back operation in the output points of the IPA, as elaborated in Appendix~\ref{appendix:ipa}.

\subsection{Iterative Update.}
The iterative update process occurs across each network layer $l$, where node features are updated, followed by edge features and frames. 
Specifically, for each layer $l$, we concatenate the updated node features $V^{l+1}_{s,i}$ and $V^{l+1}_{s,j}$.
These concatenated features undergo transformation through fully-connected layers to produce new edge features $Z^{l+1}_{s,(i,j)}$ for each time step $s$ and residue pair $(i,j)$.
Simultaneously, a frame update $\Delta T^l_i$ is computed based on the new nodes for each residue $i$ via fully-connected layers and applied to the current frame to obtain the updated frame $T^{l+1}_i$.
This iterative procedure of updating node features, edge features, and frames is repeated throughout the network, facilitating continuous propagation of updates.

\subsection{Reference Guided Motion Alignment}\label{subsec:motion}
\paragraph{Reference Network.}
The reference network is integral in encoding the structural features of the reference 3D protein structure. 
Its primary function is to ensure that the dynamic sequence generation of 3D structures retains these structural characteristics. 
Initially, we integrate the residue relationships $Z^l$ and positions $T^l$ into the node features $V^l$ for both the reference and noisy structures using the Invariant Point Attention (IPA) module. 
As illustrated in Figure~\ref{fig:spatial}, we calculate the interaction between the reference node $V^l_{\mathtt{ref}}$ and the noisy node $V^l$ by implementing a spatial module on the concatenated features $[V^l_{\mathtt{ref}},V^l_s]\in \mathbb{R}^{S \times N\times 2D}$.
For each time step $s$, the node feature is updated as follows:

\begin{equation}
    \begin{aligned}
    [A^l_{\mathtt{ref}},A^l_s] &= \mathtt{SelfAttention}( [V^l_{\mathtt{ref}},V^l_s]) \\
    \hat{V}^l_s&=A^l_sW^r + V^l_s
    \end{aligned}
\end{equation}
where $[\cdot]$ represents the collection of hidden features across dimension $D$. 
Here, $\hat{V}^l_s \in\mathbb{R}^{N\times D}$ represents the output node features for the time step $s$, and $W^r \in\mathbb{R}^{D \times D}$ is a linear projection matrix.

\paragraph{Motion Alignment.}
To accurately capture and reflect the protein's dynamic behavior, we introduce a motion alignment module. 
This component subjects the structural features of the protein to temporal self-attention within a diffusion-based generative process framework. 
Specifically, the module incorporates the 3D structures of the protein over several motion time steps preceding the reference time step, thereby embedding dynamic protein characteristics and enhancing the model’s ability to capture protein kinetics. 
We compile the node features across all time steps into a comprehensive sequence, denoted as $[V_{s}]_{s=1}^{\hat{S}}$. {$\hat{S}$ represents the total of motion $S_\texttt{mot}$, reference $S_\texttt{ref}$ and noise time steps $S$. }
Sinusoidal positional time embeddings are then added to $[V_{s}] \in \mathbb{}{R}^{N\times \hat{S} \times D}$. 
For each residue $i$, the module operates as follows:
\begin{equation}
    \begin{aligned}
    [\Tilde{A}^l_{\mathtt{mot}},\Tilde{A}^l_{\mathtt{ref}},\Tilde{A}^l]_i &= \mathtt{SelfAttention}( [V^l_s]_i) \\
    \Tilde{V}^l_i &= \Tilde{A}^l_iW^e + V^l_i
    \end{aligned}
\end{equation}
where $[\cdot]$ the collection of hidden features across the time steps $\hat{S}$. Here, $\Tilde{V}^l_i \in\mathbb{R}^{S\times D}$ denotes the output node features for time steps $\{1,...,S\}$ for residue $i$, and $W^e \in\mathbb{R}^{D \times D}$ is linear projection matrix.

\subsection{Loss Function}\label{subsec:loss}
We define the overall loss function comprising the Denoising Score Matching (DSM) loss and several auxiliary losses.

\paragraph{Denoising Score Matching Loss.} 
The neural network is trained to learn rotation and translation scores by minimizing Equation~\ref{eq:dsm}.
Specifically, we apply the weighting schedule for the rotation component as
\begin{equation}
\lambda_t^R=1/\mathbb{E}[||\nabla \log p_{t|0}(R^{(t)}|R^{(0)})||^2_{\mathrm{SO}(3)}].
\end{equation}
For the translation component, we use
\begin{equation}
\lambda_t^X=(1-\exp^{-t})/\exp^{-\frac{t}{2}}
\end{equation}
to prevent instability in loss values at low $t$.
The DSM loss is defined as follows:
\begin{equation}
\mathcal{L}_{\mathtt{dsm}}=\mathcal{L}_{\mathtt{dsm}}^{R} + \mathcal{L}_{\mathtt{dsm}}^{X}.
\end{equation}

\paragraph{Torsion Angle Loss.}
We employ a Multi-Layer Perceptron (MLP)~\cite{jumper2021highly} to predict the side chain and backbone torsion angles $\alpha_{s,i}$, represented as points on the unit circle $\Vert \alpha_{s,i} \Vert \in\mathbb{R}^{7\times 2}$ with sine and cosine values.
Due to the $180^{\circ}$ rotational symmetry of some side chains, the model is allowed to predict either the torsion angles or an alternative set of angles:
\begin{equation}
\mathcal{L}_{\mathtt{torsion}}= \frac{1}{N}\sum_{i=1}^{N} (\min (\Vert \alpha_i-\alpha_i^{gt} \Vert ^2, \Vert \alpha_i-\alpha_i^{alt gt} \Vert ^2 ))
\end{equation}
where $\alpha_i$,$\alpha_i^{gt}$ and $\alpha_i^{alt gt}$ represent predicted, ground truth, and alternative ground truth torsion angles, respectively, for each residue $i$.

\paragraph{Auxiliary loss.} 
To mitigate chain breaks or steric clashes, penalties are imposed on atomic errors.
Define $\Omega=\{\mathtt{N}, \mathtt{C}, \mathtt{C}_{\alpha}, \mathtt{O}\}$. 
The first auxiliary loss is the mean squared error on the positions of selected atoms in $\Omega$: 
\begin{equation}
\mathcal{L}_{\mathtt{\Omega}}=\frac{1}{4N}\sum_{i=1}^{N}\sum_{a\in\Omega} ||a^{(0)}_i-\hat{a}^{(0)}_i||^2
\end{equation}
where $a^(0)_i$ and $\hat{a}^(0)_i$ are the ground truth and predicted atom positions for atom $a$ in residue $i$.
The second auxiliary loss penalizes pairwise atomic distance errors:
\begin{equation}
\mathcal{L}_{\mathtt{2D}}=\frac{1}{C}\sum_{i,j=1}^N\sum_{a,b\in\Omega}\mathds{1}\{d_{ab}^{ij}<0.6\}||d_{ab}^{ij}-\hat{d}_{ab}^{ij}||^2
\end{equation}
where $C=\sum_{i,j=1}^N\sum_{a,b\in\Omega}\mathds{1}\{d_{ab}^{ij}<0.6\}-N$, $d_{ab}^{ij}=||a_{i}^{(0)}-b_{j}^{(0)}||$, and $\mathds{1}\{d_{ab}^{ij}<0.6\}$ is an indicator variable to penalize only atoms that within 0.6 nm. 
These auxiliary losses are applied only when $t<\frac{1}{4}$.

\paragraph{Total Loss.} 
The comprehensive training loss is thus formulated as:
\begin{equation}
\mathcal{L}=\mathcal{L}_{\mathtt{dsm}}+w_1\cdot \mathds{1}\{t<\frac{1}{4}\}(\mathcal{L}_{\Omega}+\mathcal{L}_{2D})+w_2\cdot \mathcal{L}_{torsion},
\end{equation}
where $w_1$ and $w_2$ are the weights for the auxiliary and torsion losses, respectively. 
In our experiments, we set $w_1=0.25$ and $w_2=1$.

\begin{figure}[t]
  \centering
  \includegraphics[width=1.0\linewidth]{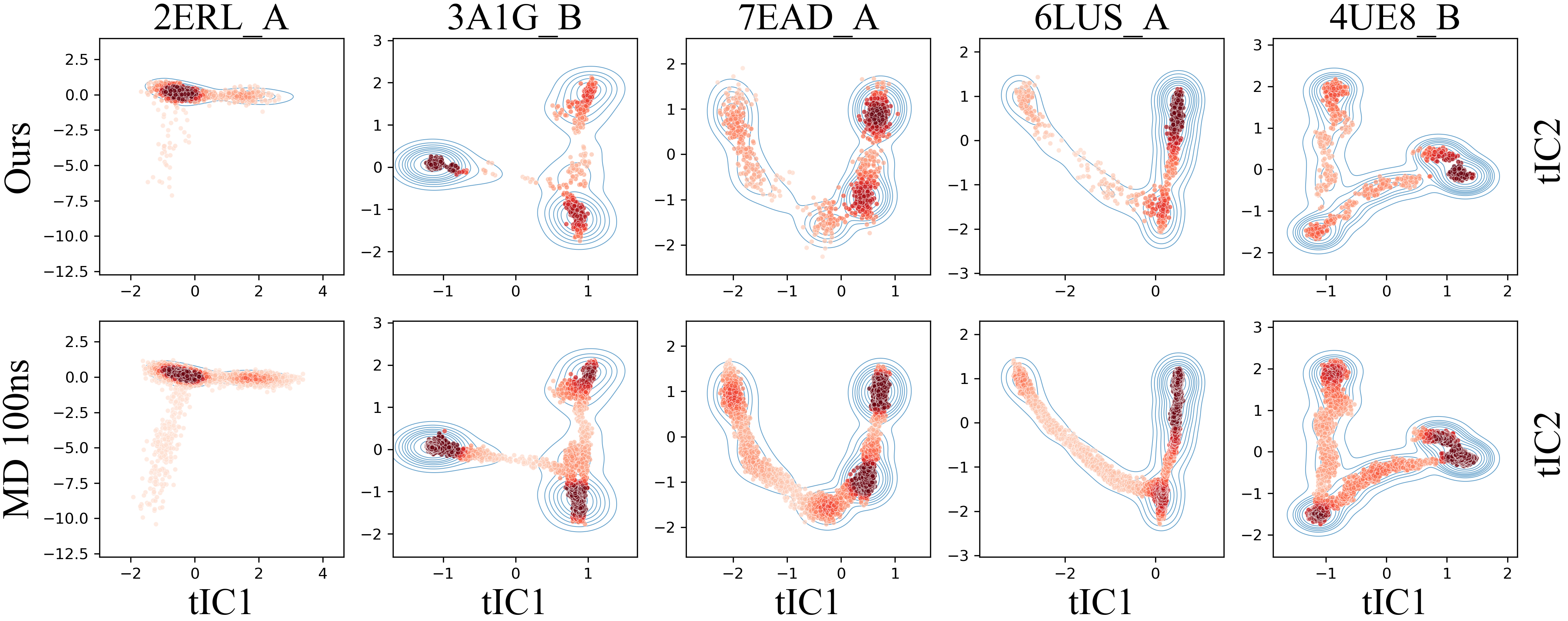}
  \caption{\textbf{Distribution Analysis.} Sample distribution over first two TIC components for different proteins. The darker the points, the higher their frequency of occurrence. The blue curve represents the kernel density distribution estimated from the MD data.}
  \label{fig:tica}
\end{figure}

\begin{figure}[t]
  \centering
  \includegraphics[width=1.0\linewidth]{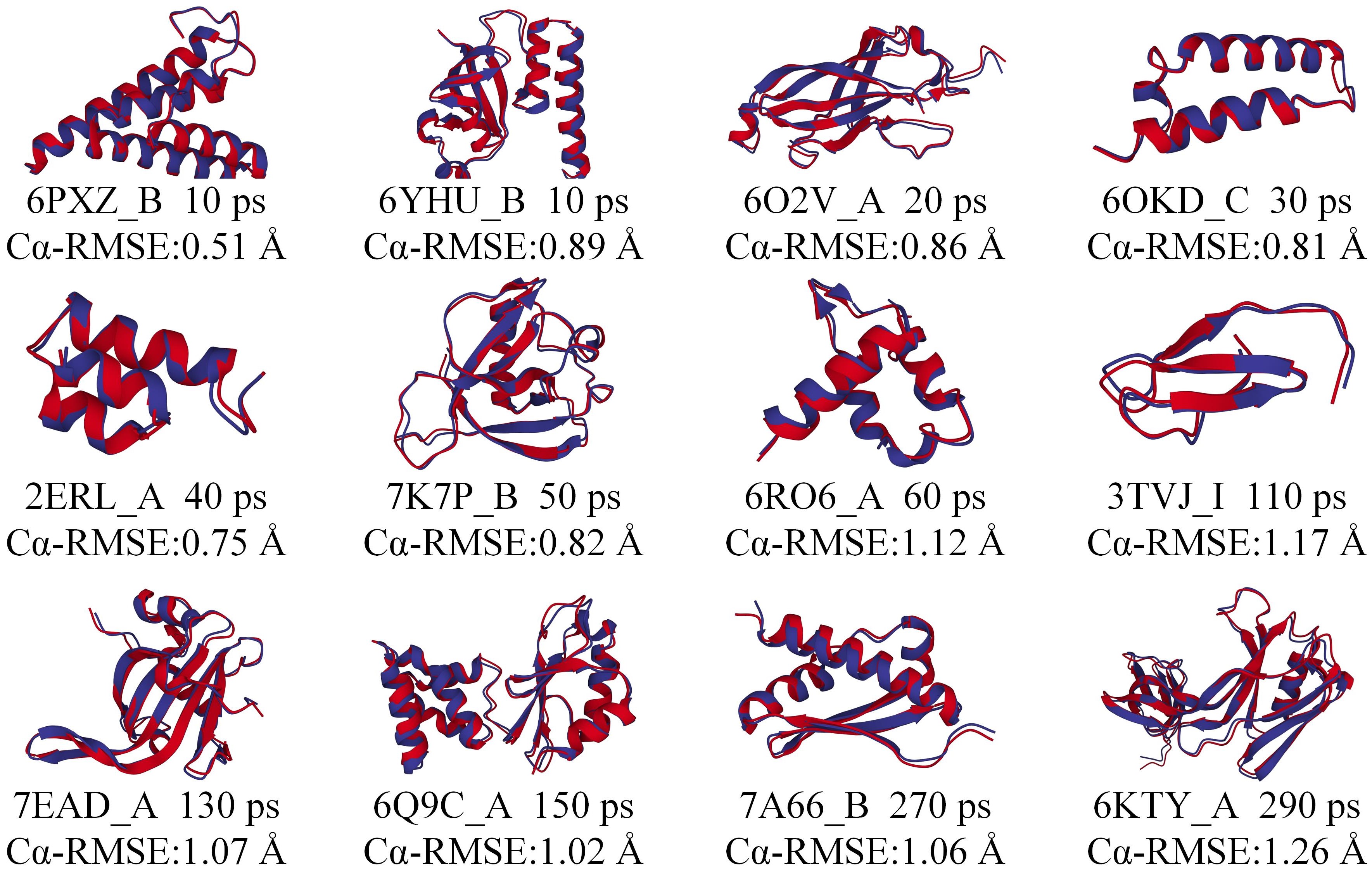}
  \caption{\textbf{Qualitative Result.} Our model prediction (blue) and the MD simulation results (red). In the first line, texts on the left refer the to protein's PDB ID and the corresponding chain, and the time on the right represents the time it takes for the reference structure to transition to this predicted structure.}  
  \label{fig:vs_gt}
\end{figure}

\begin{figure}[t]
  \centering
  \includegraphics[width=\linewidth]{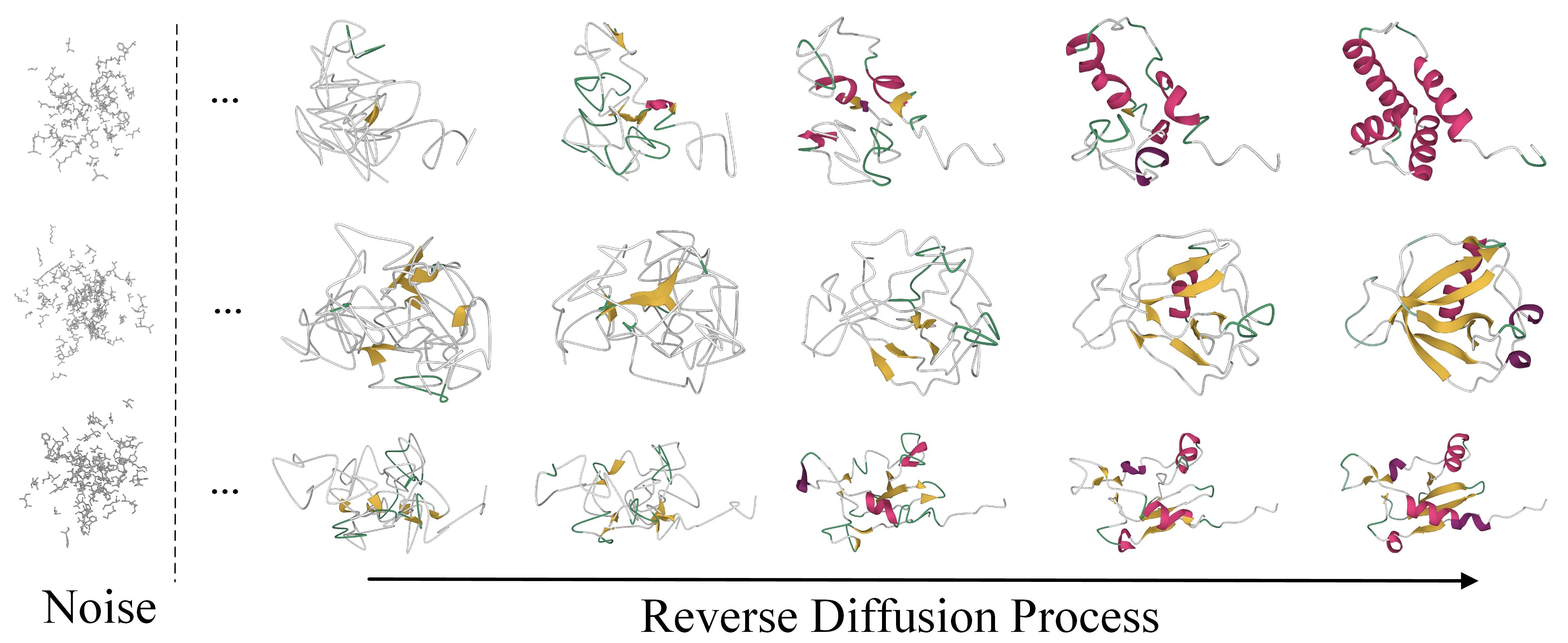}
  \caption{\textbf{ Reverse diffusion process.} {Visualization of the progression from initial noise (left) through the reverse diffusion process to form structure proteins (right). The pink and yellow regions highlight the alpha helix and beta sheets, respectively.}
  }
  \vspace{-4mm}
  \label{fig:diffprocess}
\end{figure}

\begin{figure}[t]
    \centering
    \includegraphics[width=1.0\linewidth]{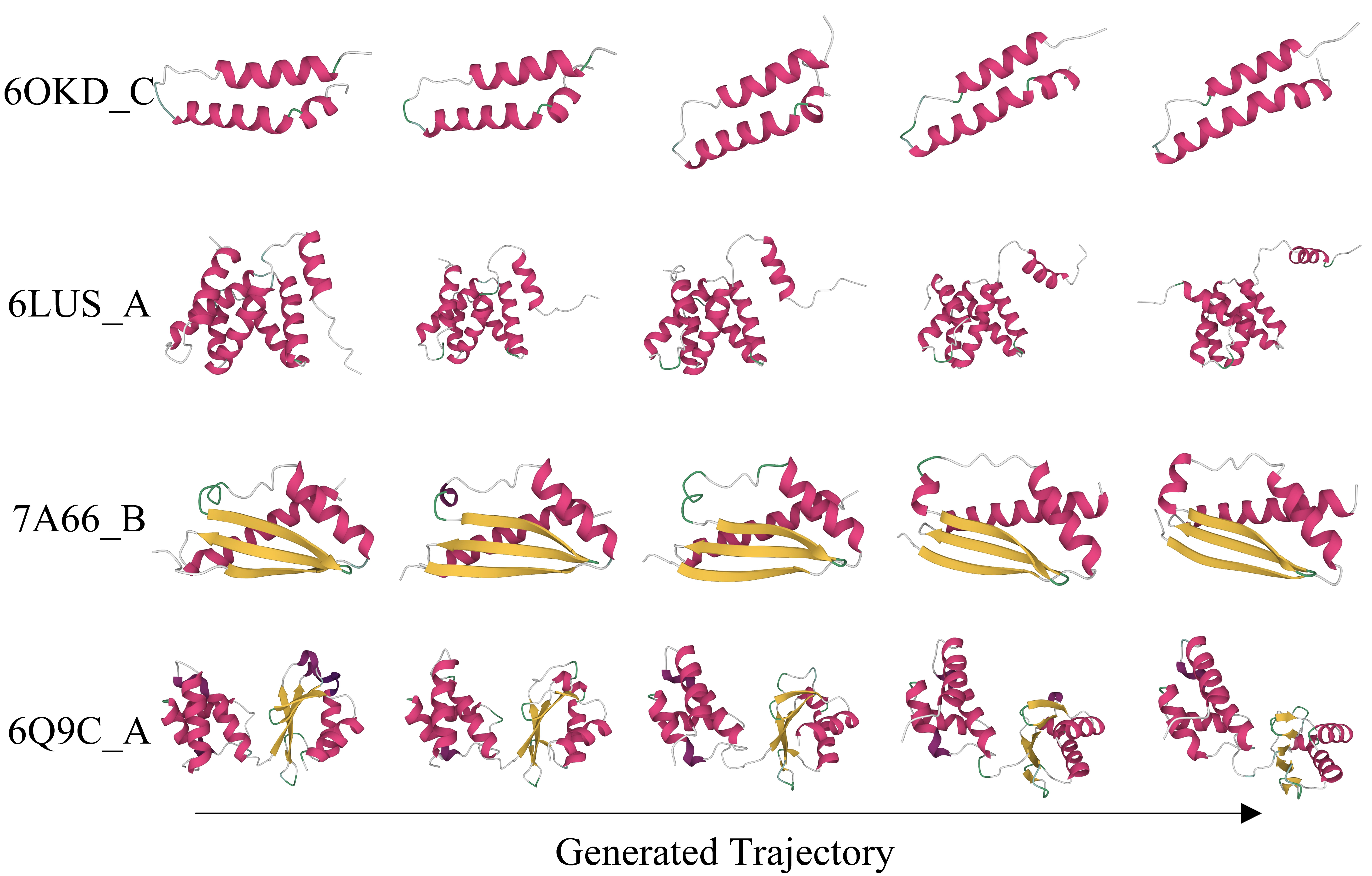}
    \caption{ \textbf{ Generated trajectories over different time steps.} The pink and yellow highlights the alpha helix and beta sheet, respectively. Our model is capable of generating conformational changes in previously unseen proteins.}
    \vspace{-4mm}
    \label{fig:vs_change}
\end{figure}

\begin{figure}[t]
    \centering
    \subfigure[Training protein numbers]{
        \begin{minipage}[b]{0.22\textwidth}
        \includegraphics[width=0.9\textwidth]{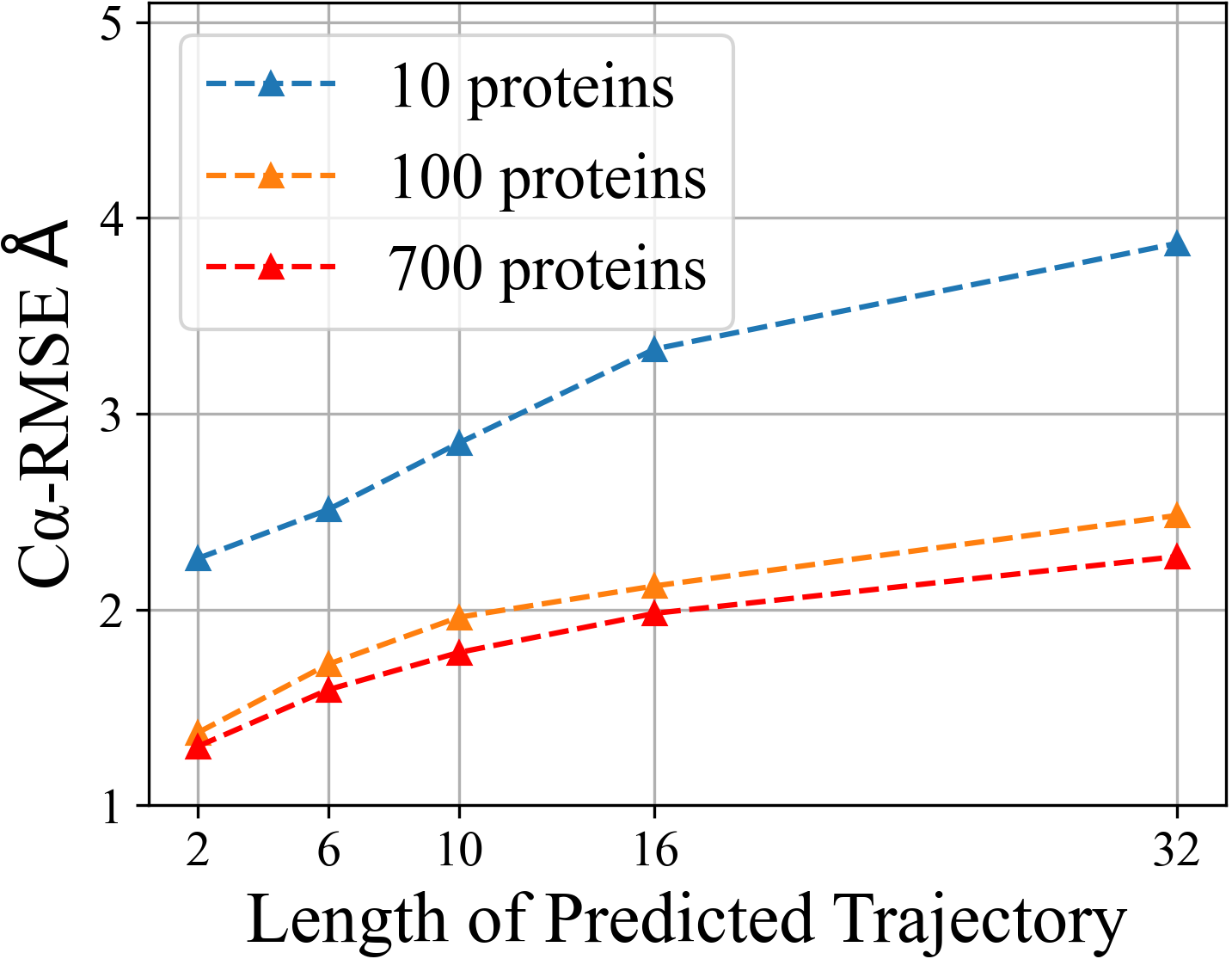}
        \label{fig:scaling_number}
        \end{minipage}
    }
    \subfigure[Training protein trajectory length]{
        \begin{minipage}[b]{0.22\textwidth}
        \includegraphics[width=0.9\textwidth]{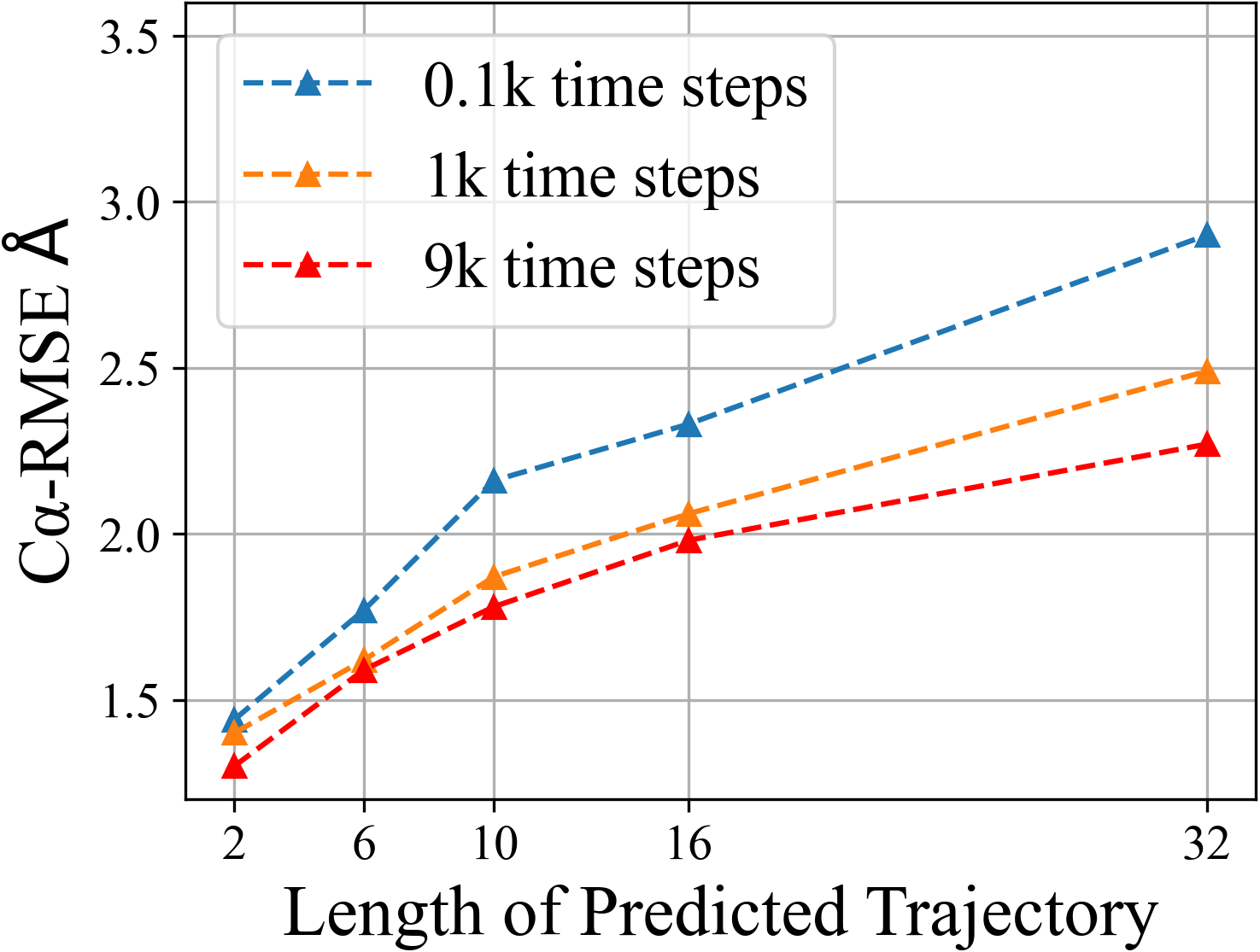}
        \label{fig:scaling_length}
        \end{minipage}
        
    }
    \caption{\textbf{Analysis of protein numbers and trajectory length on model performance}.  The performance, measured by $\rm{C}_\alpha$-RMSE, improves when a) increasing the number of training proteins or b) extending the length of training trajectories (time steps).}
    \vspace{-4mm}
    \label{fig:scaling}
\end{figure}

\section{Experiments}

% \subsection{Implementation}
% \paragraph{Dataset.} 

\paragraph{Dataset.} 
We conducted experiments and statistical comparisons against prior work utilizing datasets such as ATLAS~\cite{vander2024atlas} and fast-folding proteins~\cite{kresten2011@how}.
\textbf{(a) ATLAS}: This dataset consists of 1,390 protein chains sourced from the Protein Data Bank (PDB), selected for their structural diversity as classified by the ECOD~\cite{schaeffer2016ecod} domain classification.
To enhance the model's ability to capture structured elements like alpha-helices and beta-sheets, we used the DSSP~\cite{kabsch1983dictionary} algorithm to calculate the random coil content of each protein and excluded those with over 50\%. We also applied a polynomial regression model to filter out proteins exceeding the maximum allowable radius of gyration for their sequence length. This approach effectively removed outliers and structurally anomalous proteins, resulting in the selection of 758 proteins from the ATLAS dataset for further analysis. 
\textbf{(b) Fast-folding Proteins}: This dataset encompasses folding and unfolding events, rendering their simulated trajectories particularly complex. 
We selected six proteins, which are Chignolin, Trp\_cage, BBA, Villin, BBL, and protein\_B, for our experimental investigations.

\paragraph{Implementation Details.} 
Our framework is implemented using PyTorch 1.13.1 and Python 3.9, utilizing CUDA 11.4 for acceleration. 
All experiments and statistical analyses presented in this paper were conducted on a computing machine equipped with an NVIDIA A100 GPU with 80 GB of memory. 
We trained the parameters of our network using a batch size of 4, with an initial learning rate set to 0.0001, which subsequently decreases according to a cosine annealing schedule. 
The training procedure consists of a total of 550 epochs, with reference time steps $S_{\texttt{ref}}=1$ and motion time steps $S_{\texttt{mot}}=2$.

\subsection{Quantitative Results}

\paragraph{Task.}
Following the methodology established in DiffMD \cite{fang2023diffmd}, we empirically evaluate our approaches on two tasks:
\textbf{(a) Short-term-to-long-term (S2L) Trajectory Generation.}
In this task, models are trained on short-term trajectories and are subsequently required to generate long-term trajectories for the same protein, given a specified starting conformation. 
The training process utilizes the first 90\% of the frames, while validation and testing are conducted on the remaining 10\% of the frames. This time-based extrapolation is designed to evaluate the model's ability to generalize across the temporal view.
\textbf{(b) One-to-others (O2O) Trajectory Generation.} 
In this task, models are trained on the trajectories of a subset of proteins and evaluated on the trajectories of different proteins. 
This assessment aims to evaluate the model's ability to generalize to the conformations of distinct proteins, thereby measuring its performance across various protein types.

\paragraph{Metric.}
We adopt the Root Mean Square Error (RMSE), expressed in angstroms $\mathring {\mathrm A}$, as the evaluation metric for all snapshots over a specified time period comprising $S$ time steps, denoted as $\{s\}_{s=1}^S$. 
We define R$_s$ as the average RMSE calculated over the first $s$ time steps. 
The term $\rm{C}_\alpha$-RMSE refers to the RMSE computed between carbon alpha atoms. 
To derive our results, we sampled 500 snapshots and calculated the average RMSE across these samples.

\paragraph{Comparison to SOTA.}
In this section, we compare our framework with S2L model DFF~\cite{arts2023two} and FlowMatching~\cite{kohler2023flow}, utilizing the ATLAS and fast-folding protein datasets.
The results are summarized in Tables~\ref{tab:sota_atlas} and \ref{tab:sota_fastfolding}, where our framework demonstrates superior accuracy across both datasets. 
Notably, our approach excels in long-term predictions, as evidenced by a reduction in the R$_{32}$ error from 4.60 to 2.12 on the ATLAS dataset, and from 5.48 to 4.39 on the Fast-Folding protein dataset on the S2L task. Additionally, our model shows strong performance on the O2O task, comparable to that of S2L, underscoring its impressive generalization capability. The inclusion of proteins with longer simulation times entails greater kinetic variations at each trajectory step, further highlighting the efficacy of our method. 

\begin{table}
  \centering
  \caption{Comparison of $\rm{C}_\alpha$-RMSE among DFF, FM, and our approach on the ATLAS protein datasets.
  }
  \resizebox{\linewidth}{!}{
  \begin{tabular}{c|ccccc}
    \toprule
    %\multicolumn{2}{c}{Part}                   \\
    %\cmidrule(r){1-7}
    \multirow{2}{*}{Method} & 
    \multicolumn{5}{c}{ATLAS}  \\
    %cmidrule(r){2-11}
    & R$_2$ & R$_6$ & R$_{10}$ & R$_{16}$ & R$_{32}$  \\
    \midrule
    DFF~\cite{arts2023two}     &6.30  &7.22  &7.72 &8.11  &8.54  \\ 
    FM ~\cite{kohler2023flow}  &2.16  &2.97  &3.39 &3.82  &4.60  \\ 
    \midrule
    Ours (O2O)  & 1.30 &1.59 &1.78 & 1.98 & 2.27  \\
    Ours (S2L) & 1.27 & 1.54  & 1.67  & 1.89 & 2.12    \\
    \bottomrule
  \end{tabular}
  }
  \label{tab:sota_atlas}
\end{table}

\begin{table}
  \centering
  \caption{Comparison of $\rm{C}_\alpha$-RMSE among DFF, FM, and our approach on the FastFolding protein datasets.}
  \resizebox{\linewidth}{!}{
  \begin{tabular}{c|ccccc}
    \toprule
    %\multicolumn{2}{c}{Part}                   \\
    %\cmidrule(r){1-7}
    \multirow{2}{*}{Method} & 
    \multicolumn{5}{c}{Fast-Folding Protein} \\
    %cmidrule(r){2-11}
    & R$_2$ & R$_6$ & R$_{10}$ & R$_{16}$ & R$_{32}$ \\
    \midrule
    DFF~\cite{arts2023two}     &7.62  &7.69  &7.75 &7.83  &7.95  \\ 
    FM ~\cite{kohler2023flow}  &2.90  &3.85  &4.36 &4.84  &5.48  \\ 
    \midrule
    % Ours(generalize)      \\
    Ours (S2L)  &2.60  &3.14 &3.51 &4.06 &4.39    \\ %2.49
    \bottomrule
  \end{tabular}}
  \label{tab:sota_fastfolding}
\end{table}

\subsection{Qualitative Results} 
We visualize the distribution of dynamic proteins across the first two TIC generated by our model and compare it with the ground truth, as depicted in Figure~\ref{fig:tica}. We can see that our model effectively predicts the kinetics of proteins, aligning closely with the ground truth distribution.
The error between the predicted values (blue) and the actual MD simulation results (red) is presented in Figure~\ref{fig:vs_gt}.
The predictions maintain a $\rm{C}_\alpha$-RMSE within 2 \text{\AA} of the simulation results, demonstrating that our model accurately captures the MD simulation trajectory, particularly in light of the fact that the diameter of a carbon atom is approximately 1.4 \text{\AA}.
Figure~\ref{fig:diffprocess} illustrates the reverse diffusion process of our model at selected time steps, highlighting how the protein structure progressively attains greater coherence throughout the denoising process.
The qualitative results of different time steps are shown in Figure~\ref{fig:vs_change}. 
We can see that the proposed method effectively captures protein kinetics and generates plausible trajectories.

%%%%%%%%%%%%%%%%%%%%%%%%%%%%%%%%%%%%%%%%%%%%%%%%%%%%%%%%%%%%%%%%%%%%%

\begin{table}[t]
    \centering
    \caption{Ablation studies on the ATLAS dataset measured by $\rm{C}_\alpha$-RMSE.}
    \label{tab:ablation}
    \resizebox{\linewidth}{!}{
    \begin{tabular}{c|c|ccccc}
    \toprule
    \multirow{2}{*}{\begin{tabular}[c]{@{}c@{}}Spatial\\ Module\end{tabular}} & 
    \multirow{2}{*}{\begin{tabular}[c]{@{}c@{}}Motion\\ Alignment\end{tabular}} & 
    \multicolumn{5}{c}{ATLAS} \\  
    & & R$_2$ & R$_6$ & R$_{10}$ & R$_{16}$ & R$_{32}$ \\ 
    \midrule
    $\checkmark$ & & 1.40 & 1.66 & 1.86 & 2.07 & 2.35 \\ 
    $\checkmark$ & $\checkmark$ & 1.30 & 1.59 & 1.78 & 1.98 & 2.27 \\ 
    \bottomrule
    \end{tabular}}
\end{table}

\subsection{Ablation Studies} 
\paragraph{Effect of Motion Alignment.} 
We conducted a series of detailed ablation studies on the ATLAS dataset to assess the effectiveness of each model component.
As indicated in Table \ref{tab:ablation}, the incorporation of motion alignment results in a reduction of the R$_{2}$ error from 1.40 to 1.30 for $\rm{C}_\alpha$-RMSE. 
Our findings highlight that motion alignment is critical for the 4D dynamic protein prediction task, as it introduces essential kinetic characteristics that enhance model performance.

\paragraph{Training Protein Number.}
We investigate the impact of increasing the number of training proteins on model performance, as illustrated in Figure~\ref{fig:scaling_number}. 
The results reveal that, when the protein trajectory length is held constant, an identifiable scaling law emerges that significantly enhances model performance with the increasing number of training proteins.

\paragraph{Training Protein Trajectory Length.}
Similarly, as depicted in Figure~\ref{fig:scaling_length}, the results demonstrate that when the number of proteins is fixed and the trajectory length is increased, a corresponding scaling law emerges that effectively improves model performance.
\paragraph{Efficiency Analysis.} 
As illustrated in Table~\ref{tab:iterative}, the error of our models remains steady as the predicted trajectory length increases. In contrast, the error of the iterative method rapidly increase with longer trajectory lengths.

\begin{table}
%\scriptsize
  \caption{Comparison of $\rm{C}_\alpha$-RMSE between the iterative approach and ours on ATLAS dataset.} %\textcolor{red}{time cost}
  \centering
  \begin{tabular}{c|ccccc}
    \toprule
    %\multicolumn{2}{c}{Part}                   \\
    %\cmidrule(r){1-7}
    \multirow{2}{*}{Method}  & 
    \multicolumn{5}{c}{ATLAS } \\
    %cmidrule(r){2-11}
    & R$_2$ & R$_6$ & R$_{10}$ & R$_{16}$ & R$_{32}$  \\
    \midrule
    Iterative & 1.41 & 1.98 & 2.38 & 2.93 & 3.22 \\  
    \textbf{Ours}  & 1.30 &1.59 &1.78 & 1.98 & 2.27  \\ % ours
    % iterative
    \bottomrule
  \end{tabular}
  \label{tab:iterative}
\end{table}

\paragraph{Iterative vs. Simultaneous Prediction.} 
Our framework generates dynamic structures for $S$ time steps simultaneously. 
In contrast to previous approaches~\cite{arts2023two, fang2023diffmd}, which predict each step iteratively, we demonstrate that our design achieves superior accuracy. 
To facilitate a comparison, we also evaluate our model using an iterative approach. Specifically, we train our model to predict the 3D structure solely for the next time step, subsequently generating the entire trajectory through iterative predictions. 

\begin{figure}[t]
  \centering
  \includegraphics[width=\linewidth]{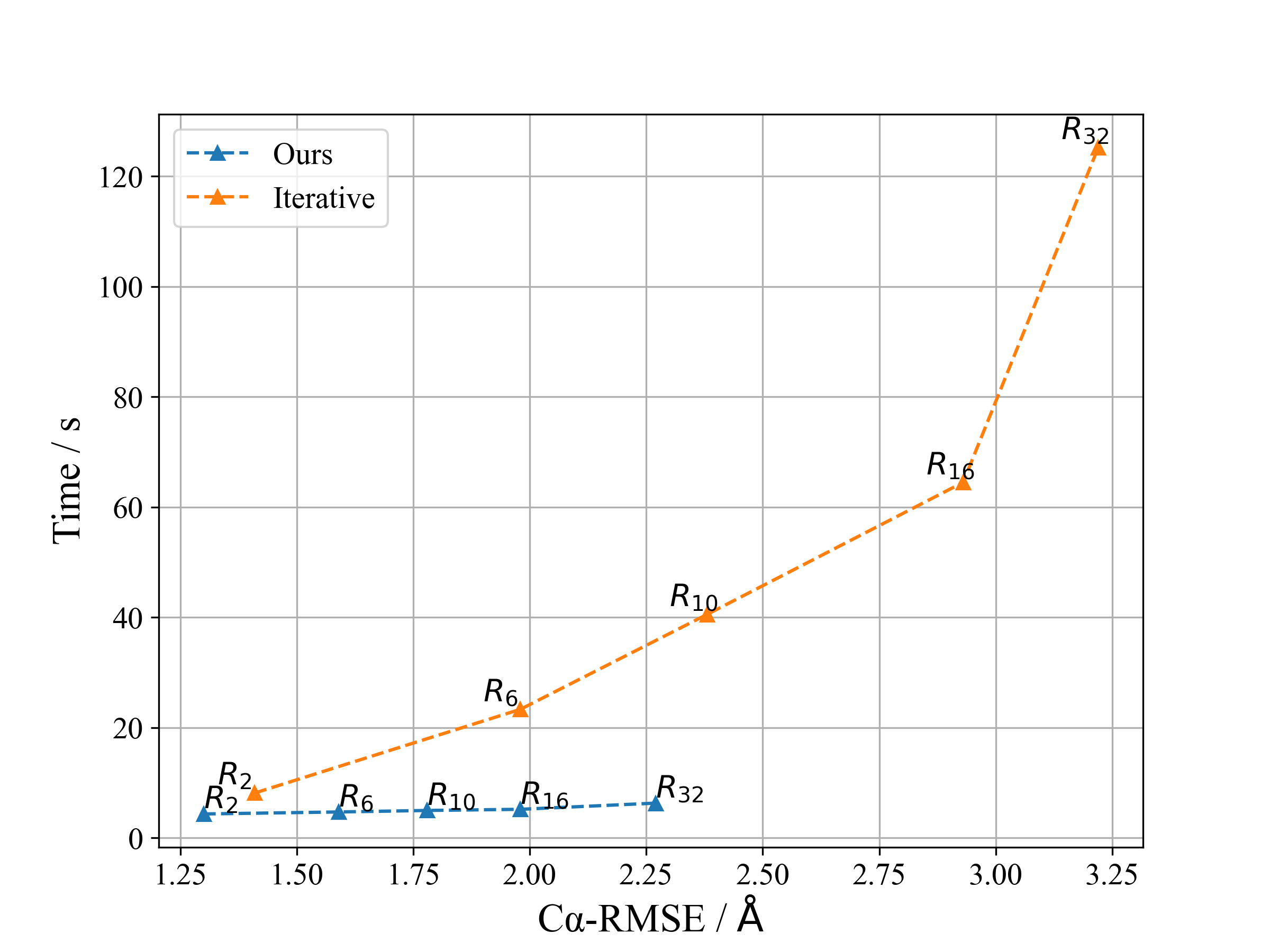}
  \caption{\textbf{Efficiency Analysis.} We visualize the error and speed curve in ATLAS. Points closer to the bottom left corner indicate higher accuracy and faster speed.}
  \label{fig:efficienty}
  \vspace{-4mm}
\end{figure}

\paragraph{Efficiency Analysis.} 
As illustrated in Figure~\ref{fig:efficienty}, the time consumption remains steady at 7 seconds as the predicted trajectory length increases. In contrast, both the time consumption and error generated by the iterative method rapidly increase with longer trajectory lengths.

\subsection{Limitations and Future Works}
Our current model effectively addresses both local movements in relatively stable states and conformational changes in proteins containing up to 256 amino acids over 32 time steps.
Additionally, it exhibits a degree of extrapolative capability, facilitating the generation of long-term molecular dynamics processes. 
Looking ahead, we intend to focus on three key areas for improvement.
(a) Longer Temporal Predictions: To enhance the accuracy of long-term predictions, we propose incorporating dynamic energy or force constraints into diffusion-based generative models. This integration will ensure that predictions remain stable and consistent with molecular dynamics principles.
(b) Improving Predictions of Large Conformational Changes: We aim to diversify the training data and expand the scale of conformational changes to bolster the model's ability to accurately predict stable, large conformational transformations.
(c) Addressing Computational Complexity: To accommodate longer amino acid sequences in protein structures, we plan to investigate the sparsity of edge feature representations, which will help alleviate computational complexity.

\section{Conclusion}
This work presents a 4D diffusion model designed to efficiently generate dynamic protein structures across multiple time steps simultaneously. 
Our unified diffusion model produces protein structures that include both the main chain and side chains. 
Additionally, we introduce a reference network that ensures structural consistency of proteins during motion, along with a motion alignment module that enhances sequential coherence in the generated dynamic proteins, thereby reducing abrupt transitions during motion. 
Our framework achieves accurate predictions by training on the ATLAS and Fast-Folding protein datasets, enabling it to explore long-term trajectories and capture feasible conformational changes.

\bibliography{aaai25}
\appendix

\setcounter{figure}{0}
\renewcommand{\thefigure}{A.\arabic{figure}}

\section{Appendix}
% This supplementary material provides additional descriptions of methodologies, dataset information, implementation details, and results that further elaborate on the experimental procedures and findings. The material is organized as follows:
% \begin{itemize}
% \item Methodology Details: IPA, EgdeUpdata, Backbone Update, Loss Function
% \item Implement Details: Training and Inference, Datasets
% \item Additional Qualitative Results: iterative results, more figures.
% \item Limitation and Future Works
% \end{itemize}

\subsection{Modules} 
\subsubsection{EdgeUpdate and BackboneUpdate} \label{appendix:edgeupdate and bbupdate}
Here we provide the detail of IPA presented in our method. Node features $V=[V_{s,i}]\in\mathbb{R}^{S\times N \times D_V}$ and edge features  $Z=[Z_{s,(i,j)}]\in\mathbb{R}^{S\times N\times N\times D_Z}$, where $V_{s,i}\in\mathbb{R}^{D_V}$ represents the feature of the residue $i$ at the time step $s$, and $Z_{s,(i,j)}\in\mathbb{R}^{D_Z}$ is to encode the relation between residues $i$ and $j$ at the time step $s$.  The transformations $T_{s,i}=[R_{s,i}, X_{s,i}]\in\mathrm{SE}(3)$, where $s\in\{1,...,S\}$, $i\in\{1,...,N\}$, $R_{s,i}\in\mathrm{SO}(3)$ is a $3\times3$ rotation matrix, and $X_{s,i}\in\mathbb{R}^3$ is the translation vector. 
\subsubsection{EdgeUpdate}
As shown in Figure~\ref{fig:network}, the edge features are updated with node features:

\begin{equation*}
\begin{array}{rl}
    \mathbf{V}_\text{down} = \text{Linear}(\mathbf{V}^{l+1}), & \mathbf{V}_\text{down} \in \mathbb{R}^{N,D_V/2} \\
    \mathbf{Z}^{\mathrm{in}}_{ij} = \mathrm{concat}(\mathbf{V}_{\mathrm{down},i}, \mathbf{V}_{\mathrm{down},j}, \mathbf{Z}^{l}_{ij}), & \mathbf{Z}^{\text{in}}_{ij} \in \mathbb{R}^{N,(D_V+D_Z)} \\
    \mathbf{Z}^{l+1} = \text{LayerNorm}(\text{MLP}(\mathbf{Z}^{\text{in}})), & \mathbf{Z}^{l+1} \in \mathbb{R}^{N,N,D_Z}
\end{array}
\end{equation*}

\subsubsection{BackboneUpdate}
For each layer $l$, we follow AlphaFold2 and update the transformation $T$ with linear projection:
\begin{equation*}
\begin{aligned}
    b_i, c_i, d_i, X_{\text{update},i} &= \mathrm{Linear}(V^{l}_i) \\
    (a_i, b_i, c_i, d_i) &= \frac{(1, b_i, c_i, d_i)}{\sqrt{1 + (b_i)^2 + (c_i)^2 + (d_i)^2}} \\
    R_{\mathrm{update},i} &= \texttt{Quat2Rot}(a_i, b_i, c_i, d_i) \\
    % \begin{pmatrix}
    %     (a_i)^2 + (b_i)^2 - (c_i)^2 - (d_i)^2 & 2b_i c_i - 2a_i d_i & 2b_i d_i + 2a_i c_i \\
    %     2b_i c_i + 2a_i d_i & (a_i)^2 - (b_i)^2 + (c_i)^2 - (d_i)^2 & 2c_i d_i - 2a_i b_i \\
    %     2b_i d_i - 2a_i c_i & 2c_i d_i + 2a_i b_i & (a_i)^2 - (b_i)^2 - (c_i)^2 + (d_i)^2
    % \end{pmatrix} \\
    T_{\text{update},i} &= (R^{\text{update}}_{i}, X^{\text{update}}_{i}) \\
    T^{l+1}_{i} &= T^{l}_{i} \cdot T^{\text{update}}_{i}
\end{aligned}
\end{equation*}
where $\texttt{Quat2Rot}(\cdot)$ represents the conversion from a quaternion to a rotation matrix.

\subsubsection{Invariant Point Attention.}\label{appendix:ipa}
To maintain the reference structures, we add features without performing the mapping back operation $T_i^{-1}$ on the output points of the IPA, as shown in line ~\ref{alg:PIA_change} of Algorithm~\ref{alg:IPA}. 

\begin{algorithm}[H]
\caption{Invariant point attention (IPA)}
\textbf{Input}: $\{\mathbf{V}_i\}$, $\{\mathbf{Z}_{ij}\}$, $\{T_i\}$, $N_{\text{head}} = 8$, $c = 256$, $N_{\text{query points}} = 8$, $N_{\text{point values}} = 12$ \\
\textbf{Output}: $\{\tilde{\mathbf{V}}_i\}$ \\
\label{alg:IPA}
\begin{algorithmic}[1]
    \STATE $\mathbf{q}_i^h, \mathbf{k}_i^h, \mathbf{v}_i^h = \text{LinearNoBias}(\mathbf{V}_i)$
    \STATE $\overrightarrow{\mathbf{q}}_i^{hp}, \overrightarrow{\mathbf{k}}_i^{hp} = \text{LinearNoBias}(\mathbf{V}_i)$
    \STATE $\overrightarrow{\mathbf{v}}_i^{hp} = \text{LinearNoBias}(\mathbf{V}_i)$
    \STATE $b_{ij}^h = \text{LinearNoBias}(\mathbf{z}_{ij})$
    \STATE $w_C = \sqrt{\frac{2}{9 N_{\text{query points}}}}$
    \STATE $w_L = \sqrt{\frac{1}{3}}$
    \STATE $a_{ij}^h =$$\text{softmax}_j \Bigg( w_L \Big( \frac{1}{\sqrt{c}} {\mathbf{q}_i^h}^T\mathbf{k}_j^{h} + b_{ij}^h \Big)$ \\
        \quad $\hfill - \frac{\gamma^h w_C}{2} \sum_p \Big\| T_i \circ \overrightarrow{\mathbf{q}}_i^{hp} - T_j \circ \overrightarrow{\mathbf{k}}_j^{hp} \Big\|^2 \Bigg)$
    \STATE $\overrightarrow{\mathbf{o}}_i^h = \sum_j a_{ij}^h \mathbf{z}_{ij}$
    \STATE $\mathbf{o}_i^h = \sum_j a_{ij}^h \mathbf{v}_j^h$
    \STATE $\overrightarrow{\mathbf{o}}_i^{hp} = T_i^{-1} \circ \sum_j a_{ij}^h \left( T_j \circ \overrightarrow{\mathbf{v}}_j^{hp} \right)$
    \STATE $\overrightarrow{\mathbf{{o}^{\prime}}}_i^{hp} =  \sum_j a_{ij}^h \left( T_j \circ \overrightarrow{\mathbf{v}}_j^{hp} \right)$  \label{alg:PIA_change}
    
    \STATE $\tilde{\mathbf{V}}_i = \text{Linear} \left( \text{concat}_{h,p}\left(\overrightarrow{\mathbf{\bar{o}}}_i^h, \mathbf{o}_i^h, \overrightarrow{\mathbf{o}}_i^{hp},\right.\right.$\\
    \quad $\hfill \left. \left. \left\| \overrightarrow{\mathbf{o}}_i^{hp} \right\|, \overrightarrow{\mathbf{{o}^{\prime}}}_i^{hp}, \left\| \overrightarrow{\mathbf{{o}^{\prime}}}_i^{hp} \right\| \right) \right)$

    \STATE \textbf{return} $\{\tilde{\mathbf{V}}_i\}$
\end{algorithmic}
\end{algorithm}
where $\overrightarrow{\mathbf{q}}_i^{hp}, \overrightarrow{\mathbf{k}}_i^{hp} \in \mathbb{R}^3, p \in \{1, \dots, N_{\text{query points}}\}$, and $\overrightarrow{\mathbf{v}}_i^{hp} \in \mathbb{R}^3, p \in \{1, \dots, N_{\text{point values}}\}$.

\begin{figure}[t]
  \centering
  \includegraphics[width=0.9\linewidth]{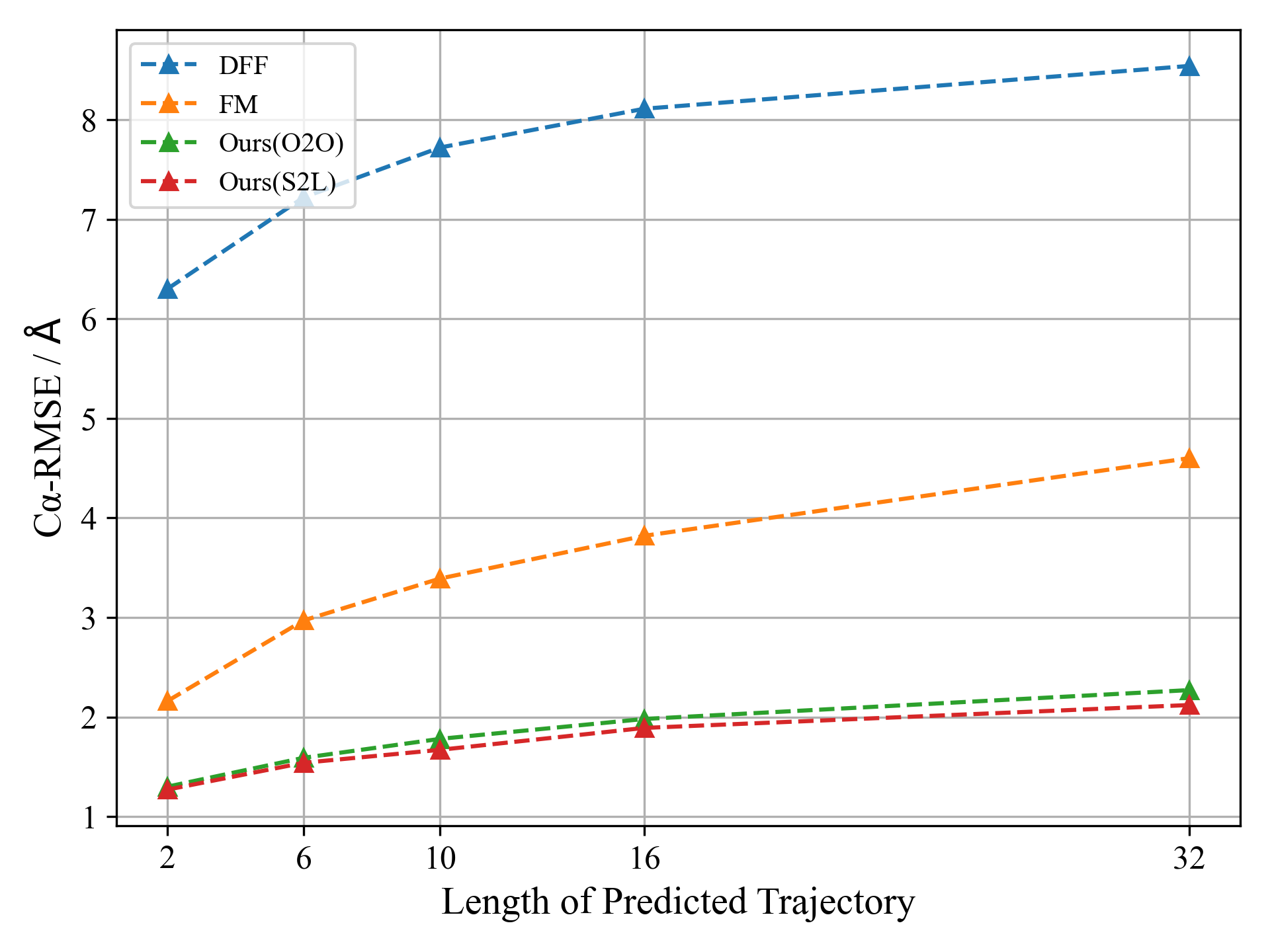}
  \caption{The error curves for different methods. Our method outperforms the others, exhibiting a lower rate of error increase as the frame rate rises.}
  \label{fig:ComparsionToSOTA}
\end{figure}

\begin{figure}[!t]
    \centering
    \includegraphics[width=1.0\linewidth]{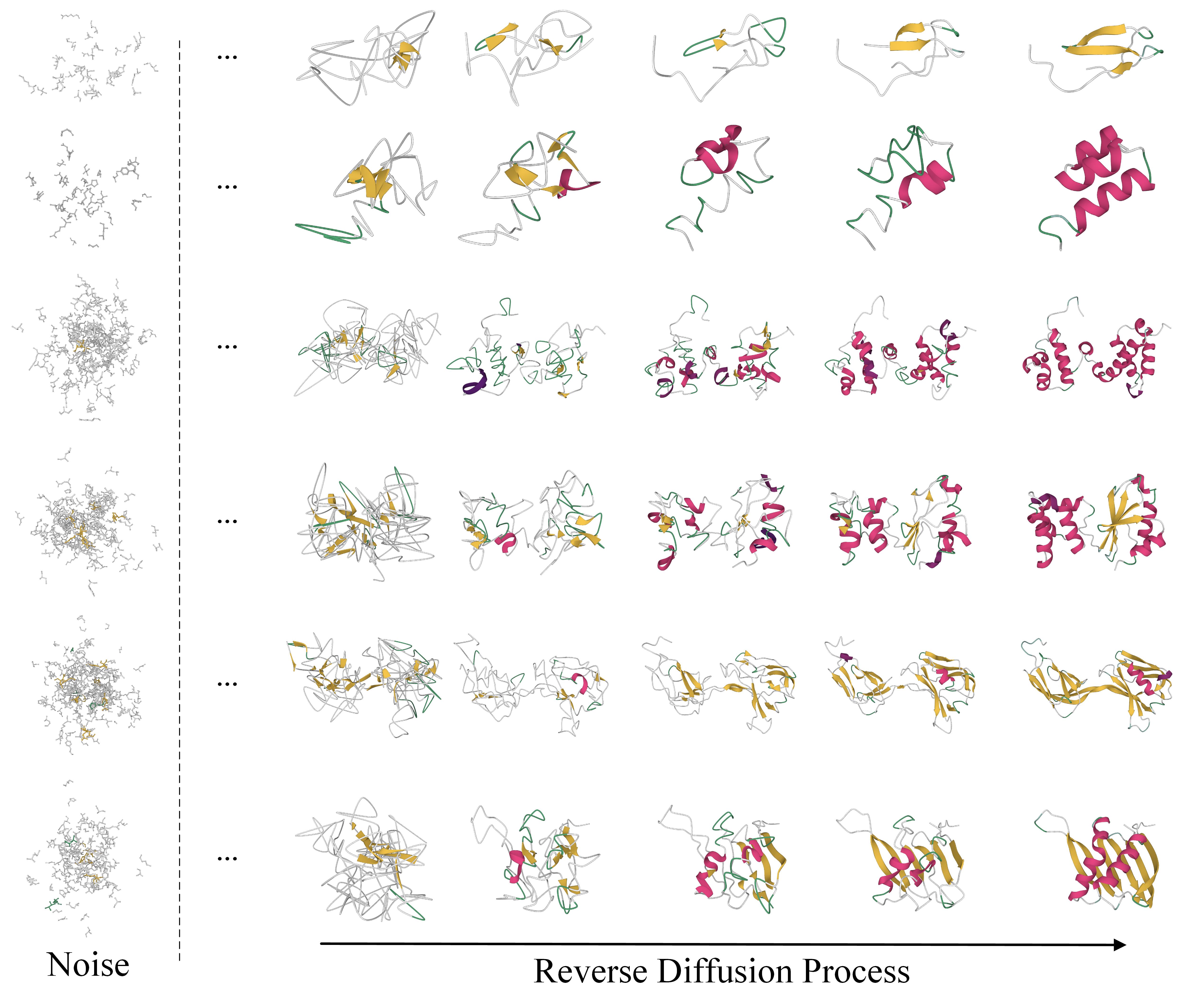}
    \caption{\textbf{ Reverse diffusion process.} {Visualization of the progression from initial noise (left) through the reverse diffusion process to form structure proteins (right). The pink and yellow regions highlight the alpha helix and beta sheets, respectively.}}
    \label{appfig:diffprocess}
    \vspace{-4mm}
\end{figure}

\subsection{Training and Inference}\label{subsec:train}
\paragraph{Training.}
The training process involves encoding the features of protein amino acid sequences through node and edge representations, alongside a corresponding noise structure that is depicted via rotations and translations, and a reference structure similarly represented. 
The output comprises denoised, continuous 3D positions of amino acids over a time segment, articulated through rotations and translations.
The procedure begins with the random sampling of fixed-length continuous time series of 3D protein structures from the original dynamic protein data, using the initial 3D structure as the reference. 
Training is conducted in two distinct stages.
In the first stage, the model learns the mapping from protein amino acid sequences to multiple 3D protein structures at various time steps. 
During this phase, the weights associated with the GeoFormer for amino acid encoding are held constant. 
In contrast, the weights responsible for the joint encoding of amino acid edges and nodes, IPA, the reference network, and updates for edges and backbones are optimized.
The second stage focuses on refining the motion alignment module to enhance the kinetic consistency of the predicted protein 3D structures within their trajectories. 
During this phase, only the parameters of the motion alignment module are optimized, while the other parameters remain unchanged.

\paragraph{Inference.}
During the inference process, the input comprises a reference 3D protein structure along with its corresponding residue sequence. 
The residue sequence is transformed into a latent representation using the GeoFormer. 
Subsequently, the denoising network employs this latent residue sequence representation, in conjunction with the reference structure, to generate the final sequences of 3D protein structures through a score-based diffusion process.

\subsection{Additional Results}

\subsubsection{Compared to SOTA.} 
Figure~\ref{fig:ComparsionToSOTA} illustrates that as the output timestep increases, our method exhibits a slower error growth rate compared to the other evaluated approaches.

\begin{figure}[!t]
    \centering
    \includegraphics[width=1.0\linewidth]{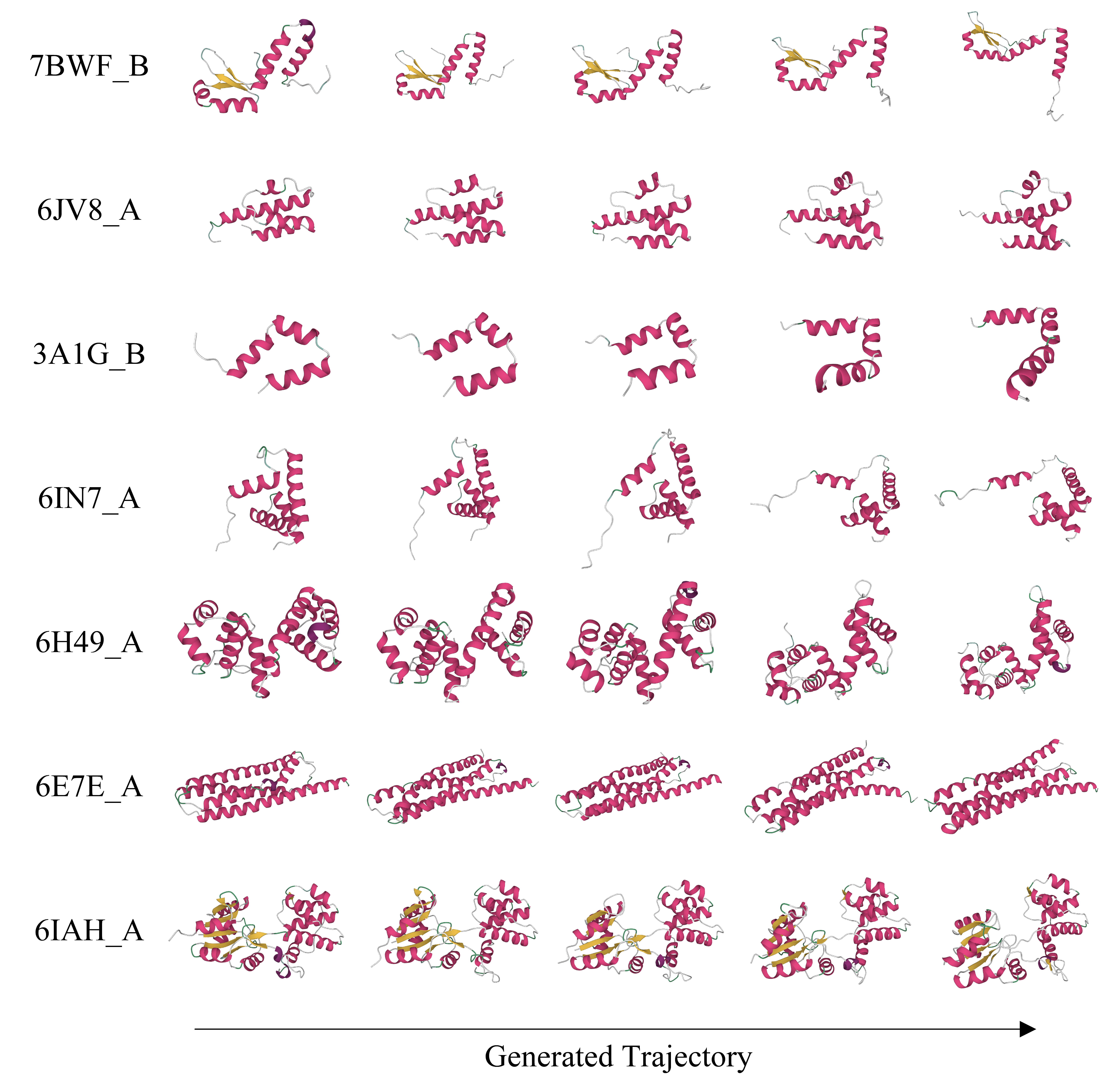}
    \caption{ \textbf{ Generated trajectories over different time steps.} The pink and yellow highlights the alpha helix and beta sheet, respectively. }
    \label{appfig:vs_change}
\end{figure}

\begin{figure}[!t]
    \centering
    \includegraphics[width=1.0\linewidth]{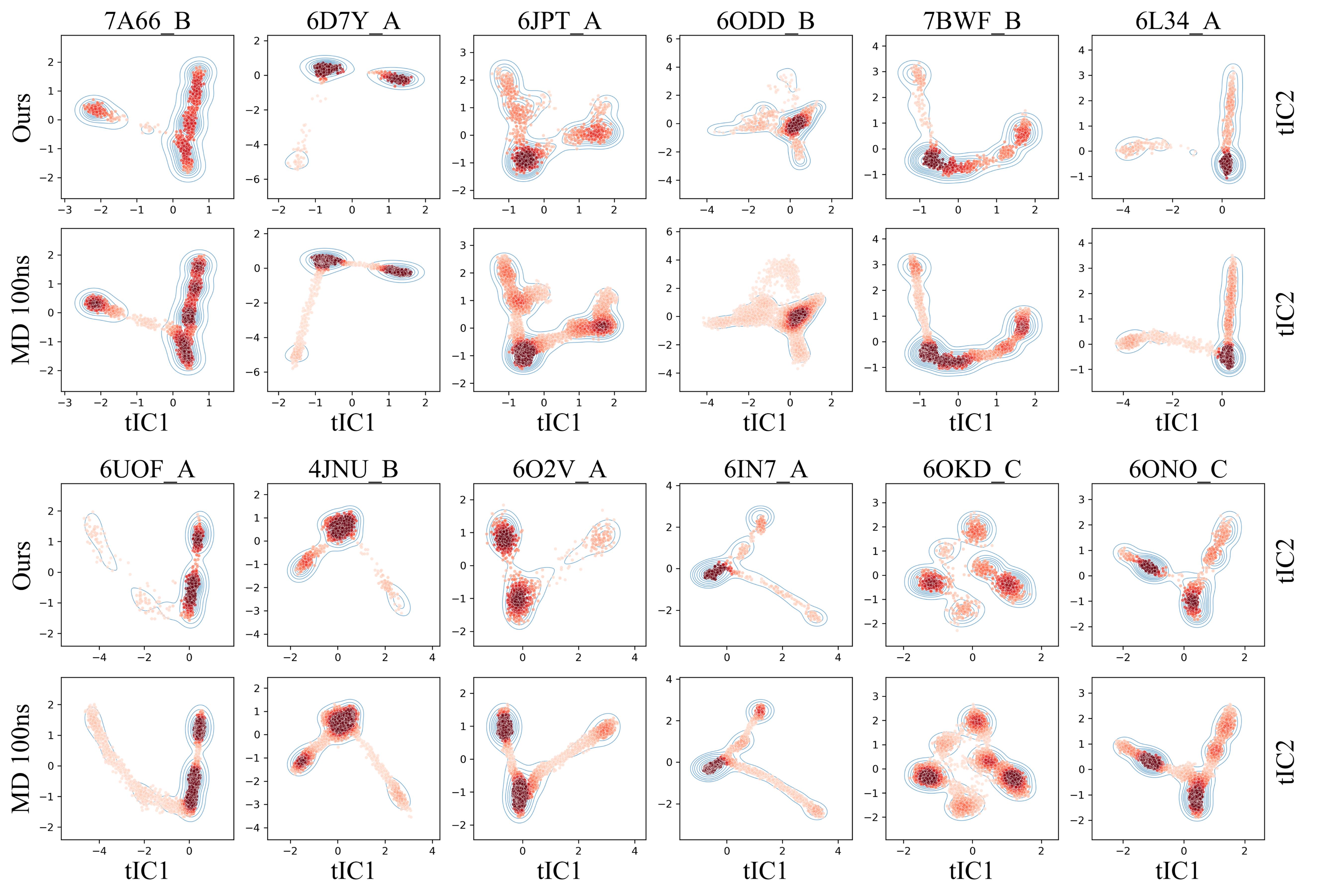}
    \caption{\textbf{Distribution Analysis.} Sample distribution over first two TIC components for different proteins. The darker the points, the higher their frequency of occurrence. The blue curve represents the kernel density distribution estimated from the MD data.}
    \vspace{-3mm}
    \label{appfig:tica}
    
\end{figure}

\subsubsection{Qualitative Results.}

Figure~\ref{appfig:diffprocess} illustrates the reverse diffusion process of our model at various selected time steps, demonstrating how the protein structure gradually becomes more refined and cohesive throughout the denoising process. We present additional results across different time steps in Figure~\ref{appfig:vs_change}. The proposed method clearly demonstrates its ability to effectively capture protein kinetics and generate realistic trajectories.
Additionally, we provide more visualization of the distribution of dynamic proteins across the first two TIC generated by our model, as shown in Figure~\ref{appfig:tica}, and compare it with the ground truth. The results demonstrate that our model accurately predicts protein kinetics, closely aligning with the ground truth distribution.

\end{document}